\documentclass{article}
\usepackage{graphicx} 

\title{Industrial Engineering with Large Language Models: A case study of ChatGPT's performance on Oil \& Gas problems }
\author{$^1$Oluwatosin Ogundare, $^2$Srinath Madasu, $^3$Nathanial Wiggins}
\date{%
    $^1$California State University, San Bernardino\\[2ex]%
    $^2$IBM Technology\\[2ex]%
    $^3$University of Houston\\[2ex]%
    \today
}

\begin{document}

\maketitle

\begin{abstract}
   Large Language Models (LLMs) have shown great potential in solving complex problems in various fields, including oil and gas engineering and other industrial engineering disciplines like factory automation, PLC programming etc. However, automatic identification of strong and weak solutions to fundamental physics equations governing several industrial processes remain a challenging task. This paper identifies the limitation of current LLM approaches, particularly ChatGPT in selected practical problems native to oil and gas engineering but not exclusively. The performance of ChatGPT  in solving complex problems in oil and gas engineering is discussed and the areas where LLMs are most effective are presented.
\end{abstract}

\section{Introduction}
In oil and gas engineering, geological models are created to represent the subsurface structure and properties of a reservoir. These models simulate fluid flow in the reservoir and are generally used to predict production performance and to optimize production strategies. Building accurate geological models are crucial for successful reservoir management and generally involve a series of  rigorous mathematical physics models. A popular technique often used in this type of problem is referred to as the Full Waveform Inversion (FWI) which involves providing accurate estimates of subsurface properties such as velocity, density  and high-resolution images of the subsurface \cite{shabelansky2007full}. The technology implication of these models is a reduction in exploration risks by optimizing well placement which leads to better recovery of the reservoir fluids \cite{naeini2016main}. Another popular problem in Oil \& Gas that also involves the analysis of waveforms is the problem of deposition and blockage detection in pipelines which uses acoustic and hydrodynamic pressure pulse reflection for pipeline inspection. The technique for pipeline inspection generally rely on the differential analysis of pressure waves or acoustic signals that are generated by a probe  or the action of a fast acting valve and then reflect back to the source after interacting with the pipe system. By analyzing the reflected waves, it is possible to detect pipeline anomalies such as corrosion, cracks, blockages or a change in cross-sectional area which can correspond to a material deposit \cite{9162624}. In the context of acoustic reflectometry and hydrodynamic echo reflection for non-intrusive pipeline inspection, the analysis of pressure waves and fluid transients can be described using potential flow and fluid transient equations which are namely the Euler and Navier-Stokes equations. These equations govern the behavior of fluid flows and can be used to model both potential flow and fluid transients, albeit with different boundary conditions that account for the presence of reflective surfaces, changes in flow direction, geometry and other engineering considerations. Closely associated with this problem is the computation of acoustic velocity in multiphase flow regimes, such as annular, bubbly, and slug, which is an important consideration with wide applications in the oil and gas industry \cite{kumar2020detecting}. Finally, for the problem of optimizing the production of an existing reservoir, a Coiled Tubing (CT) unit which typically consists of a CT reel, CT string and a control system is the principal apparatus. Generally, the Coiled Tubing (CT) unit is controlled manually by an operator using an ad hoc HMI for control when it is deployed and retrieved from the wellbore for a well intervention job, and a support team provides the necessary equipment to perform the desired intervention. The focus of recent innovations in coiled tubing operations is geared towards achieving a high degree of automation which involves the use of machine learning and specialized apparatus to minimize manual operation and enhance the efficiency and safety of well intervention jobs using an enhanced co-pilot or fully automatic self-service coiled tubing units \cite{ogundare2022tubular}\cite{ogundare2023gripper}\cite{jagnnathan2021stochastic}.
\section{Large Language Models and Mathematical Modelling}

To evaluate how a Large Language Model (LLM) like ChatGPT performs on problems related to mathematical physics modelling, it is imperative to review its analytical problem solving strategies. Wei et al. (2022) summarized that the prevailing approach adopted by Large Language Models (LLMs) is to solve mathematical problems using Chain of Thought (COT) reasoning\cite{wei2022chain}. This strategy is a step-by-step decomposition of the problem into smaller sub-problem in an iterative fashion such that every iteration step contributes towards the overall solution. One way to evaluate the performance of Large Language Models on any given mathematical problem is to check the final result. A better way is to evaluate every step in the chain of thought and assign a score to the intermediate result. We adopt the Spontaneous Quality (SQ) score to check the result at every iteration and the Reference Test (RT) score to check the final result as a measure of the composite score across all iterations\cite{ogundare2023comparative}. This is especially useful if the solution involves mixed modes, i.e., it is used within the context of a larger automatically generated control program\cite{ogundare2023no}.

For the common problem of geological prospecting, discussed in \cite{li2020coupled}\cite{kirsch2011introduction}, a reasonable task for a Large Language Model (LLM) is to formulate a mathematical model involving the general geomechanical properties like porosity, permeability, young modulus, Poisson ratio that play a role in characterizing a rock or a reservoir. The following pair of equations where generated by ChatGPT:

\begin{equation}
\frac{\partial S}{\partial t} = \frac{1}{\phi} \frac{\partial}{\partial x_i} \left( \phi \frac{k}{\mu} \frac{\partial p}{\partial x_i} \right)
\end{equation}

\begin{equation}
\rho = \rho_0 (1 - S)^{m}
\end{equation}

where S is the fluid saturation, t is time, $x_i$ represents the spatial coordinates, k is the permeability of the rock, $\mu$ is the viscosity of the fluid, p is the pressure, $\phi$ is the porosity of the rock, $\rho_0$ is the density of the rock when dry, and m is the cementation exponent. \\
When asked to formulate an inverse problem to calculate permeability, it suggested minimizing a quadratic error surface, given as follows
\begin{equation}
\mathcal{L}(k) = \left\| \rho_{obs} - \rho_{sim}(k) \right\|_{2}
\end{equation}
and selecting the value for permeability as the argument that minimizes the error or misfit function as follows:
\begin{equation}
k_{1} = \arg\min_{k} \mathcal{L}(k)
\end{equation}

All of these seem like a standard way of dealing with this type of problem. From a chain of thought perspective, if we examine the nature of the generated model using the Spontaneous Quality (SQ) score and for the overall task of generating a mathematical model, we compute the Reference Test (RT) score as the geometric mean of the SQ scores as follows:

\begin{equation}
     SQ\hspace{1mm}score = \sum_{i} \alpha_{i} P_{i}
\end{equation}
\begin{flushright}
     Where $\alpha_{i}$ = Coefficient of Importance\\
     $P_{i}$ = Performance criteria
\end{flushright}
\begin{equation}
Reference\hspace{1mm}Test (RT)\hspace{1mm}score = (\prod_{i=1}^{n} SQ_{i})^{1/n}
\end{equation}
If we select the coefficients of importance {0.5, 0.5} corresponding to accuracy and clarity, then we may have the following:\\

\begin{equation}
SQ\hspace{1mm}score\hspace{1mm}for\hspace{1mm}Equation\hspace{1mm}1: = 0.5 \times 1 + 0.25 \times 1 = 1\\
\end{equation}
The main problem with equation 2, is the introduction of the cementation exponent which is used widely when fluid saturation is a critical idea but generally not in relation to density, $\rho$. We therefore score the second equation as follows\\

\begin{equation}
SQ\hspace{1mm}score\hspace{1mm}for\hspace{1mm}Equation\hspace{1mm}2: = 0.25 \times 1 + 0.25 \times 1 = 0.5\\
\end{equation}
The idea for inversion follows the standard hill climbing or descent approaches and is assigned an SQ score of 1.\\

\begin{equation}
Reference\hspace{1mm}Test (RT)\hspace{1mm}score = (1.0 \times 0.5 \times 1.0)^{1/3} = 0.794
\end{equation}

When more precise terminology was used on another common geological problem involving single phase viscous flow in an homogeneous porous media, it generate the following continuity and momentum equations for the 2D case:
\\
Continuity equation:
\begin{equation}
\frac{\partial}{\partial x}(u(y,x)S(x)) + \frac{\partial}{\partial y}(v(y,x)S(x)) = 0
\end{equation}

\begin{equation}
\nabla \cdot (uS, vS) = 0
\end{equation}

Momentum equation:
\begin{equation}
\nabla^2 \mathbf{u} = \frac{1}{\mu} \frac{k}{\phi} \nabla p
\end{equation}
\\
where $\mathbf{u}=(u,v)$ is the velocity vector, $\nabla^2$ is the Laplacian operator, $\mu$ is the fluid viscosity, $k$ is the permeability of the porous medium, $\phi$ is the porosity of the medium, and $p$ is the fluid pressure.\\

When asked to satisfy the no-slip boundary conditions since it generated a 2D model, it generated the following:
\\ \\
If the porous medium is bounded by a rectangle with sides $x=0$, $x=L$, $y=0$, and $y=H$, we can impose the following boundary conditions:
\begin{equation}
u(0,y) = u(L,y) = 0
\end{equation}
\begin{equation}
v(x,0) = v(x,H) = 0
\end{equation}
For a numerical solution that can be solved using a standard numerical solver, ChatGPT was able to discretize the equations using a finite difference scheme.

For the continuity equation, we have:

\begin{equation}
u_{i,j+1} - u_{i,j} + v_{i+1,j} - v_{i,j} = 0
\end{equation}

For the momentum equation, we have:

\begin{equation}
u_{i+1,j} + u_{i-1,j} + u_{i,j+1} + u_{i,j-1} - 4u_{i,j} = \frac{\Delta x^2}{\mu} \frac{k}{\phi} \frac{\Delta p}{L}
\end{equation}

ChatGPT also realized that the no-slip boundary condition would be applied as follows:

\begin{equation}
    u_{0,j} = 0, u_{N,j} = 0, v_{i,0} = 0, v_{i,M} = 0
\end{equation}

where $N$ and $M$ are the number of grid points in the $x$ and $y$ directions respectively. $i$ and $j$ also ranges from 0 to $N$ and 0 to $M$ respectively.

Generally speaking, this solution would work and for the continuous case, Equations (10) through (12) would be assigned an SQ score of 1. Equations 13 \& 14, relies on geometry and ChatGPT struggles to identify the mathematical realization of the no-slip boundary conditions in complicated geometry, as such we assign an SQ score of 0.75. For the discrete case, Equation 15 is generally expected and we assign an SQ score of 1. Equation 16 assumes a uniform distribution of the pressure gradient like so $\frac{\Delta p}{L}$, which characterizes the geometry without explaining how the variation in the pressure field depends on the geometry in a rigorous way via a set of additional mathematical identities and a note that this is indeed the case. As such, we assign an SQ score of 0.5. Equation 17 also suffers the same fate as Equations 13 \& 14 and we assign an SQ score of 0.75. Consequently for the overall performance of developing model for the single phase viscous flow in an homogeneous porous media we have the RT score below:
\begin{equation}
Reference\hspace{1mm}Test (RT)\hspace{1mm}score = (\prod_{i=1}^{n} SQ_{i})^{1/n} = 0.8534
\end{equation}
\section{Improving the performance of LLMs on Oil \& Gas problems}
Despite the outstanding performance of Large Language Models (LLMs) on the reference problems identified in this paper, there is still a significant deficit in its performance on the type of problems that make up the frontier of Oil \& Gas research efforts, especially the application of machine learning to automatic control of complicated apparatus in exploratory and production operations. On the surface, the quality of the data might be identified as a major reason for the dearth of creativity in the solutions generated by LLMs on the Oil \& Gas and other related problems, but it appears that this notion might not be fundamentally true as shown in the solutions presented in this paper. The two solutions generated by ChatGPT had a strong theoretical basis and were fairly rigorous. These ideas presented by ChatGPT were consistent with what is expected to be learned at the graduated level by experts working on the problem. What appears to be missing is the ability to extrapolate the theory to unusual scenarios. Some of these scenarios are not necessarily more complex than the what is established in the knowledge space of a LLM but rather a semantic parallel with equivalent complexity. For example on the 2D model of viscous flow in homogeneous porous media, suppose the porous medium was bounded by a trapezoid, how does ChatGPT respond to the interpretation of the no-slip boundary condition? It maintains the same interpretation of the boundary conditions which doesn't reflect the change in the geometry as follows:
\\
$u(0,y) = u(L,y) = 0$ (no flow through the sides)\\
$v(x,0) = v(x,H) = 0$ (no flow through the top and bottom)
\\
This necessarily holds as it encapsulates every shape into an enclosing rectangle and insists that the no-slip condition holds within that region. For instance, a simple improvement would be removing the assumption that $x$ is fixed through the sides with an expression 
\begin{equation}
    u(x, \frac{x}{cos(\theta)}) = 0
\end{equation}
Which will hold for an arbitrary value of $\theta$ that is bounded such that the resulting geometry remains a trapezoid and values of $x$ such that $\frac{\partial y}{\partial x} \neq 0$. This of course is not severe but demonstrates the type of extrapolation you would expect from a model that aspires to work on real world problems. The complexity of the change is also trivial and simply locates a slanted line when the values $\theta$ are correctly bounded and an easy improvement to incorporate into a production LLM model.
\\ \\
\section{Performance of LLMs on Analytical PDEs in Oil \& Gas}
Consider the 1D case of Burger's equation that describe inviscid flow in an ideal oil \& gas pipeline:
\begin{equation}
\frac{\partial u}{\partial t} + u\frac{\partial u}{\partial x} = 0
\end{equation}
Subject to the following initial condition $u(x, 0) = 4x(1-x),  0 < x < 1$ and homogeneous boundary conditions- $u(0,t) = u(1,t) = 0$ and $u(x,t) = 2\Pi v\frac{\sum_{n=1}^{\infty}a_{n}e^{-n^{2}\pi^{2}vt}nSin(n \pi x)}{a_{0} + \sum_{n=1}^{\infty}a_{n}e^{-n^{2}\pi^{2}vt}Cos(n \pi x) }$ where $a_{0}$ and $a_{n}$ are Fourier coefficients given by $\int_{0}^{1} exp{-(2\pi v)^{-1}[1 - Cos(\pi x)]} \,dx$ \& $2\int_{0}^{1} exp{-(2\pi v)^{-1}[1 - Cos(\pi x)]Cosn(\pi x)} \,dx$ for $n = 1, ..., 10$ used by Hayati et al.\cite{hayati2007feedforward}\\ \\
ChatGPT uses Chaint of Thought (COT) reasoning to decompose the problem as follows:
\begin{equation}
u(x,t) = \frac{2}{\sqrt{\pi}} \sum_{n=1}^{\infty} (-1)^{n+1} e^{-(2n-1)^2 \pi^2 v t} \frac{\sin((2n-1)\pi x)}{2n-1}
\end{equation}
The Fourier coefficients are given by:
\begin{equation}
a_n = \int_{0}^{1} \sqrt{\frac{2}{\pi}} u(x,0) \cos(n\pi x) dx
\end{equation}
where $u(x,0) = 4x(1-x)$ is the initial condition. 

Therefore, the solution for $u(x,t)$ can be written as:
\begin{equation}
u(x,t) = \frac{32}{\pi^3} \sum_{n=1}^{\infty} \frac{(-1)^{n+1}}{(2n-1)^3} e^{-(2n-1)^2 \pi^2 v t} \sin((2n-1)\pi x)
\end{equation}
For $u(0.25,t)$ where $t= 0.6, 0.8, 1.0,$ and $3.0$ we have the following results calculated by ChatGPT and paired by the expected exact solution:
\begin{equation}
u(0.25,0.6) = 0.006006\hspace{2mm} |\hspace{2mm}Exact \hspace{1mm} Solution: 0.00189
\end{equation}
\begin{equation}
u(0.25,0.8) = 0.003671 \hspace{2mm} | \hspace{2mm} Exact  \hspace{1mm}Solution: 0.00026
\end{equation}
\begin{equation}
u(0.25,1.0) = 0.002528 \hspace{2mm} |\hspace{2mm}  Exact  \hspace{1mm}Solution: 0.00004
\end{equation}
\begin{equation}
u(0.25,3.0) = 0.000067 \hspace{2mm} |\hspace{2mm}  Exact  \hspace{1mm}Solution: 0.00000
\end{equation}

Clearly, ChatGPT is good at Chain of Thought (COT) reasoning towards a solution but bad at computations that are not trivial. This is a major drawback to complete industrial adoption of LLMs in industrial automation and systems governed by non-trivial mathematical physics.

Additionally, since smaller industrial engineering organizations particularly upstarts in the Oil \& Gas servicing might find the cost of training a specialized cluster of LLMs on new techniques and methods prohibitive, a way to improve results might be a developing a domain specific software that enriches the solutions generated by a generalized LLM with new information only in the aspects where the upstarts maintain a leading competency before presenting a final output.

\section{Conclusion}

Large Language Models (LLMs) have a potential to be useful in industrial engineering, particularly oil and gas engineering. This paper has identified some of the limitations of current LLM approaches, particularly ChatGPT, in solving problems in the oil and gas industry. The potential applications of LLMs in solving problems in various areas of oil and gas engineering was demonstrated with examples from rock physics but generally includes other important aspects like Full Waveform Inversion (FWI), acoustic reflectometry, hydrodynamic pressure pulse reflection, and well intervention. Finally, areas for improvement were suggested, including improving the nature of the data used to train LLMs, enriching the output of LLMs with domain-specific knowledge which in many cases involves imposing physical constraints on the output.

\bibliographystyle{plain} 
\bibliography{refs} 

\begin{thebibliography}{10}

\bibitem{hayati2007feedforward}
Mohsen Hayati and Behnam Karami.
\newblock Feedforward neural network for solving partial differential
  equations.
\newblock {\em Journal of Applied Sciences}, 7(19):2812--2817, 2007.

\bibitem{jagnnathan2021stochastic}
Srinivasan Jagnnathan, Oluwatosin Ogundare, Srinath Madasu, and Keshava
  Rangarajan.
\newblock Stochastic realization of parameter inversion in physics-based
  empirical models, June~3 2021.
\newblock US Patent App. 16/956,605.

\bibitem{kirsch2011introduction}
Andreas Kirsch et~al.
\newblock {\em An introduction to the mathematical theory of inverse problems},
  volume 120.
\newblock Springer, 2011.

\bibitem{kumar2020detecting}
Amod Kumar, Claudio Olmi, Oluwatosin Ogundare, Pranab Jha, David Bennett,
  et~al.
\newblock Detecting pipeline anomalies and variations in acoustic velocity in
  multiphase flow regimes using computational fluid dynamics.
\newblock {\em Open Journal of Fluid Dynamics}, 10(03):184, 2020.

\bibitem{li2020coupled}
Dongzhuo Li, Kailai Xu, Jerry~M Harris, and Eric Darve.
\newblock Coupled time-lapse full-waveform inversion for subsurface flow
  problems using intrusive automatic differentiation.
\newblock {\em Water Resources Research}, 56(8):e2019WR027032, 2020.

\bibitem{naeini2016main}
Ehsan~Zabihi Naeini, Tariq Alkhalifah, Ilya Tsvankin, Nishant Kamath, and
  Jiubing Cheng.
\newblock Main components of full-waveform inversion for reservoir
  characterization.
\newblock {\em First Break}, 34(11), 2016.

\bibitem{ogundare2023comparative}
Oluwatosin Ogundare and Gustavo~Quiros Araya.
\newblock Comparative analysis of chatgpt and the evolution of language models.
\newblock {\em arXiv preprint arXiv:2304.02468}, 2023.

\bibitem{ogundare2023no}
Oluwatosin Ogundare, Gustavo~Quiros Araya, and Yassine Qamsane.
\newblock No code ai: Automatic generation of function block diagrams from
  documentation and associated heuristic for context-aware ml algorithm
  training.
\newblock {\em arXiv preprint arXiv:2304.04117}, 2023.

\bibitem{ogundare2022tubular}
Oluwatosin Ogundare and Samuel Fagbemi.
\newblock Tubular lockup prediction in deviated wells using markov chains.
\newblock In {\em International Conference on Offshore Mechanics and Arctic
  Engineering}, volume 86328, page V001T02A002. American Society of Mechanical
  Engineers, 2022.

\bibitem{9162624}
Oluwatosin Ogundare and Srinivasan Jagannathan.
\newblock Computational acoustic model for non-intrusive inspection of a
  fluidic channel.
\newblock In {\em 2020 5th Asia-Pacific Conference on Intelligent Robot Systems
  (ACIRS)}, pages 129--132, 2020.

\bibitem{ogundare2023gripper}
Oluwatosin Ogundare, Radovan Rolovic, Roshan~Joy Jacob, Ramalakshmi
  Somaskandan, Jeremy~C Nicholson, and Glenn~Marvin Ozburn.
\newblock Gripper control in a coiled tubing system, January~31 2023.
\newblock US Patent 11,566,479.

\bibitem{shabelansky2007full}
Andrey~Hanan Shabelansky.
\newblock {\em Full wave inversion}.
\newblock University of Tel-Aviv, 2007.

\bibitem{wei2022chain}
Jason Wei, Xuezhi Wang, Dale Schuurmans, Maarten Bosma, Ed~Chi, Quoc Le, and
  Denny Zhou.
\newblock Chain of thought prompting elicits reasoning in large language
  models.
\newblock {\em arXiv preprint arXiv:2201.11903}, 2022.

\end{thebibliography}
\end{document}